\documentclass[10pt,twocolumn,letterpaper]{article}

\pdfoutput=1

\usepackage{cvpr}
\usepackage{times}
\usepackage{epsfig}
\usepackage{graphicx}
\usepackage{amsmath}
\usepackage{amssymb,amsfonts,mathtools}
\usepackage{color}
\usepackage[table]{xcolor}
\usepackage{bbm}

\renewcommand{\Re}[1]{\mbox{$\mathbbm{R}^{#1}$}}

\usepackage[breaklinks=true,bookmarks=false]{hyperref}

\cvprfinalcopy 

\def\cvprPaperID{****} 

\begin{document}

\title{FERAtt: Facial Expression Recognition with Attention Net}

\author{Pedro D. Marrero Fernandez, \: Fidel A. Guerrero Pe\~{n}a, \: Tsang Ing Ren\\
Centro de Inform\'atica, Universidade Federal de Pernambuco, Brazil\\
{\tt\small \{pdmf, fagp, tir\}@cin.ufpe.br}
\and
\\
Alexandre Cunha\\
Center for Advanced Methods in Biological Image Analysis\\ California Institute of Technology, USA\\
{\tt\small cunha@caltech.edu}}

\maketitle


\begin{abstract}
We present a new end-to-end network architecture for facial expression recognition with an attention model. It focuses attention in the human face and uses a Gaussian space representation for expression recognition. We devise this architecture based on two fundamental complementary components: (1) facial image correction and attention and (2) facial expression representation and classification. The first component uses an encoder-decoder style network and a convolutional feature extractor that are pixel-wise multiplied to obtain a feature attention map. The second component is responsible for obtaining an embedded representation and classification of the facial expression. We propose a loss function that creates a Gaussian structure on the representation space. To demonstrate the proposed method, we create two larger and more comprehensive synthetic datasets using the traditional BU3DFE and CK+ facial datasets. We compared results with the PreActResNet18 baseline. Our experiments on these datasets have shown the superiority of our approach in recognizing facial expressions. 
\end{abstract}
\section{Introduction}

Human beings are able to express and recognize emotions as a way to communicate an inner state. Facial expression is the main form to convey this information and its understanding has transformed the treatment of emotions by the scientific community. Traditionally, scientists assumed that people have internal mechanisms comprising a small set of emotional reactions ({\it e.g.} happiness, anger, sadness, fear, disgust) that are measurable and objective. 
Understanding these mental states from facial and body cues is a fundamental human trait, and such aptitude is vital in our daily communications and social interactions. In fields such as human-computer interaction (HCI), neuroscience, and computer vision, scientists have conducted extensive research to understand human emotions. Some of these studies aspire to creating computers that can understand and respond to human emotions and to our general behavior, potentially leading to seamless beneficial interactions between humans and computers. Our work aims to contribute to this effort, more specifically in the area of Facial Expression Recognition, or FER for short.

\begin{figure}[t!]
\begin{center}
   \includegraphics[width=0.48\linewidth]{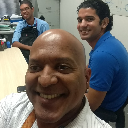}
   \hfill
   \includegraphics[width=0.49\linewidth]{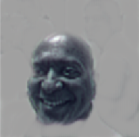}
\end{center}
\vspace{-2mm}
   \caption{Example of attention in a selfie image. Facial expression is recognized on the front face which is separated from the less prominent components of the image by our approach. Our goal is to jointly train for attention and classification where faces are segmented and their expressions learned by a dual--branch network. By focusing attention on the face features we try to eliminate a detrimental influence possibly present on the other elements in the image during facial expression classification. A differential of our formulation is thus that we explicitly target learning expressions solely on learned faces and not on other irrelevant parts of the image.}
\label{fig:att_component}
{\color{gray}\hrule}
\vspace{-4mm}
\end{figure}
Deep Convolutional Neural Networks (CNN) have recently shown excellent performance in a wide variety of image classification tasks \cite{NIPS2012_4824, Russakovsky2015, Szegedy_2015_CVPR, SimonyanZ14aDBLP}. The careful design of local to global feature learning with convolution, pooling, and layered architecture produces a rich visual representation, making CNN a powerful tool for facial expression recognition \cite{li2018deep}. Research challenges such as the Emotion Recognition in the Wild (EmotiW) series\footnote{https://sites.google.com/view/emotiw2018} and Kaggle's Facial Expression Recognition Challenge\footnote{https://www.kaggle.com/c/challenges-in-representation-learning-facial-expression-recognition-challenge} revealed the growing interest of the community in the use of deep learning for the solution of this problem, a trend we adopt in this work.

Recent developments for the FER problem consider processing the entire image regardless of the face location within the image, exposing them to potentially harmful noise and artifacts and incurring in unnecessary additional computational cost. This is problematic as the {\it minutiae} that characterize facial expressions can be affected by environmental elements such as hair, jewelry, and other objects proximal to the face but in the image background. Some methods use heuristics to decrease the searching size of the facial regions. 
Such approaches contrast to our understanding of the human visual perception, which quickly parses the field of view, discards irrelevant information, and then focus the main processing on a specific target region of interest -- the so called {\em visual attention} mechanism \cite{itti2001}. Our approach tries to mimic this behavior as it aims to suppress the contribution of surrounding deterrent elements and it concentrates recognition solely on facial regions. \figurename~\ref{fig:att_component} illustrates how the attention mechanism works in a typical scene.

Attention mechanisms have recently been explored in a wide variety of contexts \cite{NIPS2015_5635, NIPS2015_5854}, often providing new capabilities to algorithms \cite{graves2016hybrid, gregor2015draw, NIPS2016_6230}. While they improve efficiency \cite{NIPS2014_5542} and performance on state-of-the-art machine learning benchmarks \cite{NIPS2015_5635}, their computational architecture is much simpler than those comprising the mechanisms in the human visual cortex \cite{dayan2003theoretical}. Attention has also been long studied by neuroscientists \cite{ungerleider2000mechanisms}, who believe it is crucial for visual perception and cognition \cite{cheung2016emergence} as it is inherently tied to the architecture of the visual cortex and can affect its information. 

Our contributions are summarized as follows: (1) To the best of our knowledge, this is the first CNN-based method using attention to jointly solve for representation and classification in FER problems; (2) We propose a dual-branch network to extract an attention map which in turn improves the learning of kernels specific to facial expression; (3) A new loss function is formulated for obtaining a facial manifold represented as a Gaussian Mixture Model; and (4) We create a synthetic generator to render face expressions.

\section{Related Works}

Tang \cite{tang2013deep} proposed jointly learning a deep CNN with a linear Support Vector Machine (SVM) output. His method achieved the first place on both public (validation) and private data on the FER-2013 Challenge \cite{goodfellow2013challenges}. 
Liu {\it et al.} \cite{liu2014facial} proposed a facial expression recognition framework using 3DCNN together with deformable action parts constraints to jointly localize facial action parts and learn part-based representations for expression recognition. Liu \textit{et al.} \cite{liu2014combining} followed by including the pre-trained Caffe CNN models to extract image-level features. 
In the work of Kahou \textit{et al.} \cite{Kahou2013} a CNN was trained for video recognition and a deep Restricted Boltzmann Machine (RBM) was trained for for audio recognition. “Bag of mouth” features were also extracted to further improve the performance.

Yu and Zhang achieved state-of-the-art results in EmotiW in 2015 using CNNs. They used an ensemble of CNNs each with five convolutional layers \cite{yu2015image} and showed that randomly perturbing the input images yielded a 2-3\% boost in accuracy. Specifically, they applied transformations to the input images at training time. At testing time, their model generated predictions for multiple perturbations of each test example and voted on the class label to produce a final answer. Also of interest in this work is that they used stochastic pooling \cite{graham2014fractional} rather than max pooling due to its good performance on limited training data.
Mollahosseini {\it et al.} have also obtained state of the art results \cite{mollahosseini2016going} with their network consisting of two convolutional layers, max-pooling, and four inception layers, the latter introduced by GoogLeNet. Their architecture was tested on many publicly available data sets.

\section{Methodology}

In this section, we describe our contributions in designing a new network architecture, the formulation of the loss functions used for training, and our method to generate synthetic data.

\subsection{Network architecture}

\begin{figure*}
\begin{center}
\includegraphics[width=\linewidth]{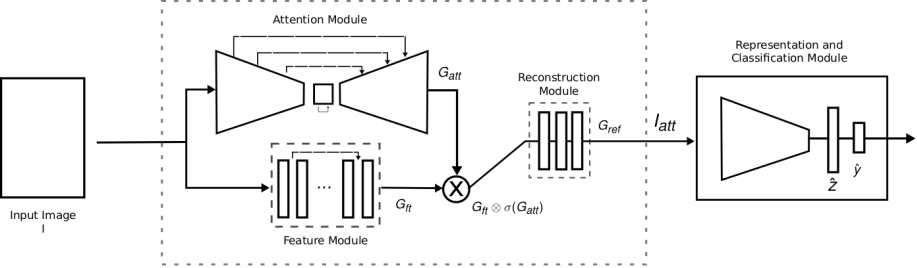}
\end{center}
  \caption{\textbf{Architecture of FERAtt}. Our model consists of four major modules: attention module $G_{att}$, feature extraction module $G_{ft}$, reconstruction module $G_{rec}$, and classification and representation module $G_{rep}$. The features extracted by $G_{att}$, $G_{ft}$ and $G_{rec}$ are used to create the attention map $I_{att}$ which in turn is fed into $G_{rep}$ to create a representation of the image. Input images $I$ have $128\times128$ pixels and are reduced to $32\times32$ by an Averaging Pooling layer on the reconstruction module. Classification is thus done on these smaller but richer representations of the original image.}
\label{fig:arquitecture}
\vspace{1mm}
{\color{gray}\hrule}
\end{figure*}

Given a facial expression image $I$, our objective is to obtain a good representation and classification of $I$. The proposed model, Facial Expression Recognition with Attention Net (FERAtt), is based on the dual-branch architecture \cite{he2017mask, li2016deep, pan2018learning, zhu2016deep} and consists of four major modules: (i) an attention module $G_{att}$ to extract the attention feature map, (ii) a feature extraction module $G_{ft}$ to obtain essential features from the input image $I$, (iii) a reconstruction module $G_{rec}$ to estimate a good attention image $I_{att}$, and (iv) a representation module $G_{rep}$ that is responsible for the representation and classification of the facial expression image. An overview of the proposed model is illustrated in \figurename ~ \ref{fig:arquitecture}.

\textbf{Attention module.} We use an encoder-decoder style network, which has been shown to produce good results for many generative \cite{shocher2018zero,zhu2016deep} and segmentation tasks \cite{RFB15a}. In particular, we choose a variation of the fully convolutional model proposed in \cite{RFB15a} for semantic segmentation. We add four layers in the coder with skip connections and dilation of 2x. The decoder layer is initialized with pre-trained ResNet34 \cite{resnetDBLP} layers. This significantly accelerates the convergence. We denote the output features of the decoder by $G_{att}$, which will be used to determine the attention feature map.

\begin{figure}[t!]
\begin{center}
\includegraphics[width=0.4\linewidth]{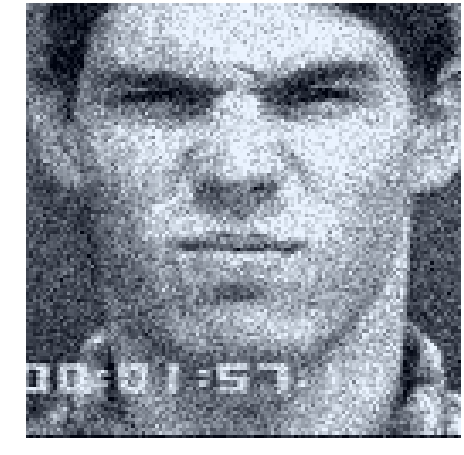} 
\includegraphics[width=0.4\linewidth]{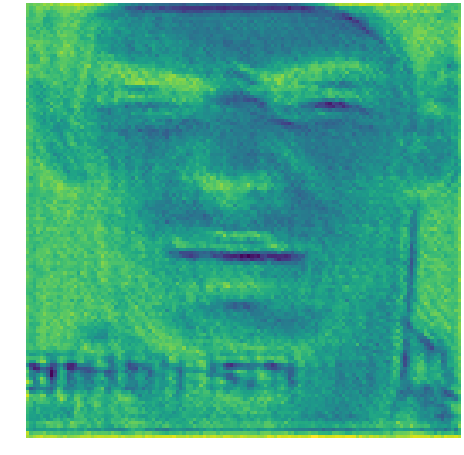}
\begin{tabular}{p{0.4\linewidth} c p{0.4\linewidth}}
 (a) Input image $I$ & (b) $G_{ft}$
\end{tabular}
\includegraphics[width=0.4\linewidth]{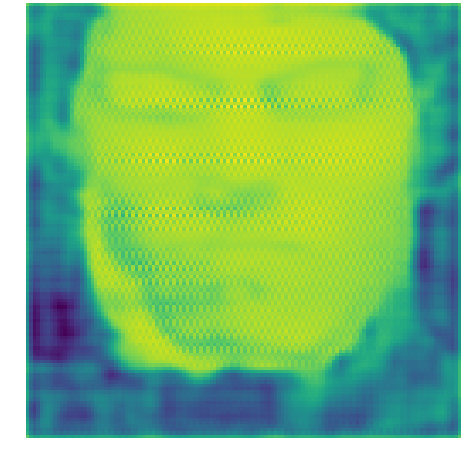}
\includegraphics[width=0.4\linewidth]{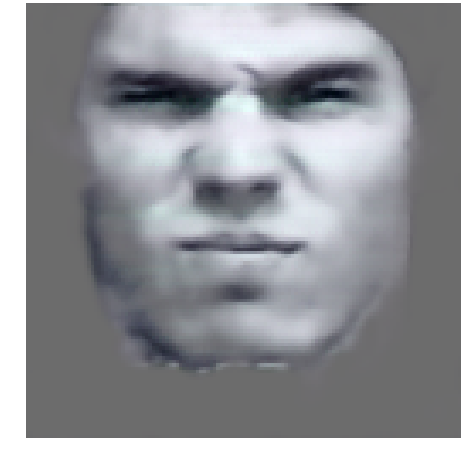}
\begin{tabular}{p{0.4\linewidth} c p{0.4\linewidth}}
(c) $G_{att}$  & (d) $I_{att}$
\end{tabular}
\end{center}
   \caption{Generation of attention map $I_{att}$. A $128\times128$ noisy input image (a) is processed by the feature extraction $G_{ft}$ and attention $G_{att}$ modules whose results, shown, respectively, in panels (b) and (c), are combined and then fed into the reconstruction module $G_{rec}$. This in turn produces a clean and focused attention map $I_{att}$, shown on panel (d), that will then be classified by the last module $G_{rep}$ of FERAtt. The $I_{att}$ image shown here is before reduction to $32\times32$ size.}
\label{fig:att_component_out}
\vspace{1mm}
{\color{gray}\hrule}
\end{figure}

\textbf{Feature extraction module.} We use four ResBlocks \cite{lim2017enhanced} to extract high-dimensional features for image attention. To maintain spatial information, we do not use any pooling or strided convolutional layers. We denote the extracted features as $G_{ft}$ -- see \figurename~\ref{fig:att_component_out}b.

\textbf{Reconstruction module.} The reconstruction layer adjusts the attention map to create an enhanced input to the representation module. It has two convolutional layers, a Relu layer, and an Average Pooling layer which, by our design choice, resizes the input image of $128\times128$ to $32\times32$. This reduced size was chosen for the input of the representation and classification module (PreActivationResNet \cite{he2016identity}), a number we borrowed from the literature and to facilitate comparisons. We plan to experiment with other sizes in the future. We denote the feature attention map as $I_{att}$ -- see \figurename~\ref{fig:att_component_out}d.

\textbf{Representation and classification module.} For the representation and classification of facial expressions, we have chosen a Fully Convolutional Network (FCN) of PreActivateResNet \cite{he2016identity}. This architecture has shown excellent results when applied on classification tasks. The output of this FCN, $z=f_{\Theta}(I_{att})$, is evaluated in a linear layer to obtain a vector $\hat{z}$ with the desired dimensions. Finally, vector $\hat{z}$ is evaluated in a regression layer to estimate the probability $p(w|\hat{z})$ for each class $w_j$.

\subsection{Loss functions}

The FERAtt network generates three outputs: a feature attention map $\hat{I}_{att}$, a representation vector $\hat{z}$, and a classification vector $\hat{y}$. In our training data, each image $I$ has an associated binary ground truth mask $I_{mask}$ corresponding to a face in the image and its expression class $y$. We train the network by jointly optimizing the sum of attention, representation, and classification losses:

\begin{equation}\label{eq:loss}
\min_{\Theta}\; \{  \mathcal{L}_{att} ( I_{att}, I \otimes I_{mask}  )  + \mathcal{L}_{rep}( \hat{z}, y ) + \mathcal{L}_{cls}( \hat{y}, y )  \}
\end{equation}
%
We use the pixel-wise MSE loss function for $\mathcal{L}_{att}$, and for $\mathcal{L}_{cls}$ we use the BCE loss function. We propose a new loss function $\mathcal{L}_{rep}$ for the representation. 

\subsection{Structured Gaussian Manifold Loss}

Suppose that we separate a collection of samples per class in an embedded space so that we have $c$ sets, $C_1, \ldots, C_c$, with the samples in $C_j = \{f_{\Theta}(x^j_1), \ldots, f_{\Theta}(x^j_{n_j}) \}$, $j = 1,2, \ldots, c$, one for each class $w_j$, and the neural net function $f_{\Theta}:\Re{D}\to\Re{d}$, are drawn independently according to probability $p(x|w_j)$ for input $x$. 

We assume that $p(f_{\Theta}(x)|w_j)$ has a known parametric form, and is therefore determined uniquely by the value of a parameter vector $\theta_j$. For example, we might have $p(f_{\Theta}(x)|w_j) \sim N(\mu_j, \Sigma_j)$, where $\theta_j = (\mu_j,\Sigma_j)$, for $N$ the normal distribution with mean $\mu_j$ and variance $\Sigma_j$. To show the dependence of $p(f_{\Theta}(x)|w_j)$ on $\theta_j$ explicitly, we write $p(f_{\Theta}(x)|w_j)$ as $p(f_{\Theta}(x)|w_j,\theta_j)$. Our problem is to use the information provided by the training samples to obtain a good transformation function $f_{\Theta}(x_j)$ that generate embedded spaces with known distribution associated with each category. Then the {\it a posteriori} probability $P(w_j|f_{\Theta}(x))$ can be computed from $p(f_{\Theta}(x)|w_j)$ by the Bayes' formula:

\begin{equation}
P(w_j| f_{\Theta}(x) ) = \frac{p(w_j)p(f_{\Theta}(x)|w_j, \theta_i)}{ \sum_i^c p(w_i)p(f_{\Theta}(x)|w_i, \theta_i) }
\end{equation}

In this work, we are using the normal density function $p(x|w_j,\theta_j)$. The objective is to generate embedded sub-spaces with a defined structure. For our first approach we use Gaussian structures:

\begin{equation}                                      
p(f_{\Theta}(x)|w_j,\mu_j,\Sigma_j) = \frac{1}{(2\pi)^{n/2}|\Sigma_j|^{1/2} } \exp( -\frac{1}{2} X^T \Sigma_j^{-1} X )    
\end{equation}
where $X=(f_{\Theta}(x)-\mu_j)$. For the case $\Sigma_j = \sigma^2I$:
\begin{equation}
p(x|w_j,\mu_j,\sigma_j) = \frac{1}{ \sqrt[]{(2\pi)^n} \sigma_j } \exp( -\frac{||f_{\Theta}(x)-\mu_j||^2}{2\sigma_j^2} )   
\end{equation}

In a supervised problem, we know the {\it a posteriori} probability $P(w_j|x)$ for the input set. From this, we can define our structured loss function as the mean square error between the {\it a posteriori} probability of the input set and the {\it a posteriori} probability estimated for the embedded space:

\begin{equation} \label{eq:gd}
\mathcal{L}_{rep} = \mathbb{E} \left \{ ||P(w_j|f_{\Theta}(x_k)) - P(w_j|x_k)||^2_2 \right \} 
\end{equation}

\subsection{Synthetic image generator}

\begin{figure}[t]
\begin{center}
\begin{tabular}{l}
\includegraphics[width=0.3\linewidth]{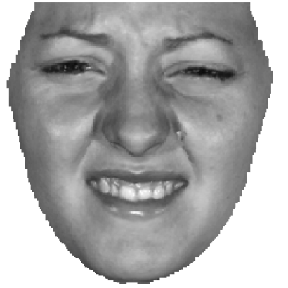} 
\includegraphics[width=0.3\linewidth]{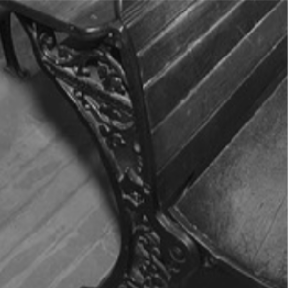} 
\includegraphics[width=0.3\linewidth]{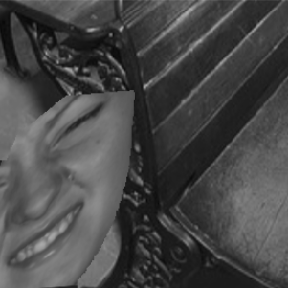} \\
~~(a) Face image ~~ (b) Background ~~ (c) Composition 
\end{tabular}
\end{center}
   \caption{Example of synthetic image generation. A cropped face image and a general background image are combined to generate a composite image. By using distinct background images for every face image we are able to generate a much larger training data set.}
\label{fig:synthetic}
\vspace{1mm}
{\color{gray}\hrule}
\end{figure}

One of the limiting problems for FER is the small amount of correctly labeled data. In this work, we propose a renderer $R$ for the creation of a synthetic larger dataset from real datasets as presented in \cite{fernandez2018fast}. $R$ allows us to make background changes and geometric transformations of the face image. \figurename~\ref{fig:synthetic} shows an image generated from an example face of the BU3DFE dataset and a background image.

The generator method is limited to making low-level features that represent small variations in the facial expression space for the classification component. However, it allows creating a good number of examples to train our end-to-end system, having a larger contribution to the attention component. In future works we plan to include high-level features using GAN from the generated masks \cite{huang2017dyadgan}.

The renderer $R$ adjusts the illumination of the face image so that it is inserted in the scene more realistically. An alpha matte step is applied in the construction of the final composite image of face and background. The luminance channel of the image face model $I_{face}$ is adjusted by multiplying it by the factor $\frac{I_{r}}{I_{face}}$ where $I_{r}$ is the luminance of the region that contains the face in the original image.

\section{Experiments}

We describe here the creation of the dataset used for training our network and its implementation details.  We discuss two groups of experimental results: (1) Expression recognition result, to measure the performance of the method regarding the relevance of the attention module and the proposed loss function, and (2) Correction result, to analyze the robustness to noise.

\subsection{Datasets}

To evaluate our method, we used two public facial expression datasets, namely Extended Cohn-Kanade (CK+) \cite{lucey2010extended} and BU-3DFE \cite{yin20063d}. In all experiments,  person-independent FER scenarios are used \cite{zeng2009survey}. Subjects in the training set are completely different from the subjects in the test set, i.e., the subjects used for training  are not used for testing. The CK+ dataset includes 593 image sequences from 123 subjects. From these, we selected 325 sequences of 118 subjects, which meet the criteria for one of the seven emotions \cite{lucey2010extended}. The selected 325 sequences consist of 45 Angry, 18 Contempt, 58 Disgust, 25 Fear, 69 Happy, 28 Sadness and 82 Surprise \cite{lucey2010extended}. In the neutral face case, we selected the first frame of the sequence of 33 random selected subjects. The BU-3DFE dataset is known to be challenging and difficult mainly due to a variety of ethnic/racial ancestries and expression intensity \cite{yin20063d}. A total of 600 expressive face images (1 intensity x 6 expressions x 100 subjects) and 100 neutral face images, one for each subject, were used \cite{yin20063d}. 

\begin{figure}[t]
\begin{center}
   \includegraphics[width=0.95\linewidth]{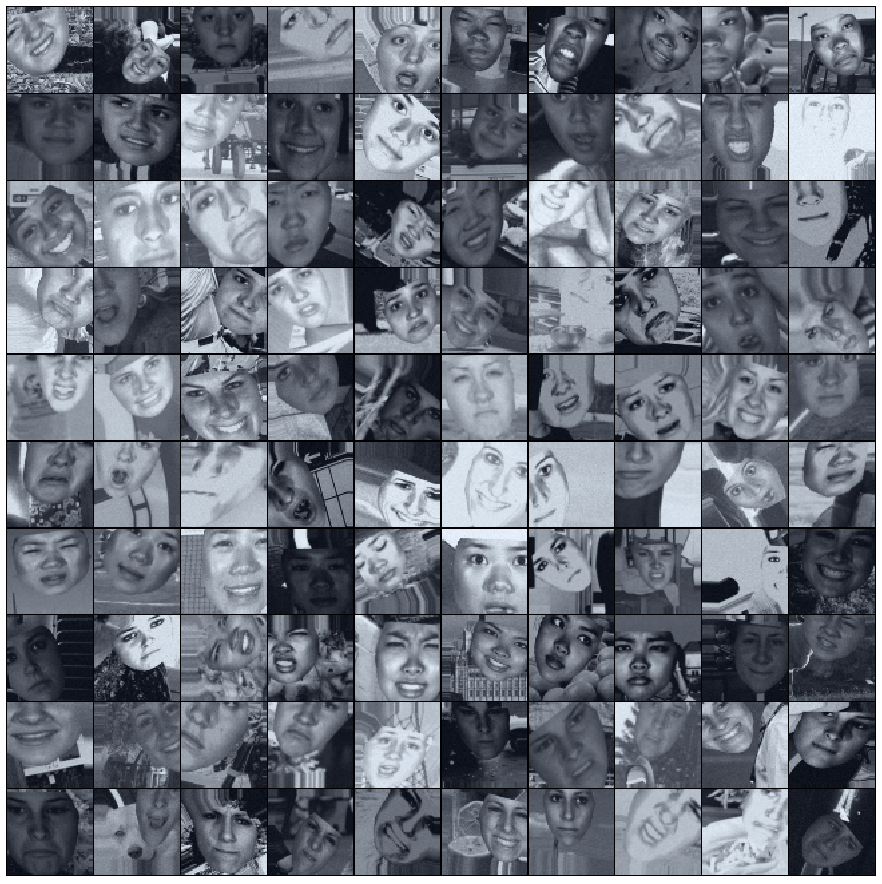}
\end{center}
   \caption{Examples from the synthetic BU-3DFE dataset. Different faces are transformed and combined with randomly selected background images from the COCO dataset. After transformation, color augmentation is apply (brightness, contrast, Gaussian blur and noise).  }
\label{fig:synthetic_dataset_bu3dfe}
\vspace{1mm}
{\color{lightgray}\hrule}
\end{figure}

We employed a renderer $R$ to create training data for the neural network. $R$ uses a facial expression dataset (we use BU-3DFE and CK+, which were segmented to obtain face masks) and a dataset of background images (we have chosen the COCO dataset). \figurename~\ref{fig:synthetic_dataset_bu3dfe} show examples of images generated by the renderer on BU-3DFE dataset.

\subsection{Implementation and training details}

In all experiments we considered the architecture PreActResNet18 for the classification and representation processes. We adopted two approaches: (1) a model with attention and classification, FERAtt+Cls, and (2) a model with attention, classification, and representation, FERAtt+Rep+Cls. These were compared with the classification results. For the representation, the last convolutional layer of PreActResNet is evaluated by a linear layer to generate a vector of selected size. We have opted for 64 dimensions for the representation vector $\hat{z}$.

All models were trained on Nvidia GPUs (P100, K80, Titan XP) using PyTorch\footnote{http://pytorch.org/} for 60 epochs on the training set with 200 examples per mini batch and employing the Adam optimizer. Face images were rescaled to 32$\times$32 pixels. The code for the FERAtt is available in a public repository\footnote{\url{https://github.com/pedrodiamel/ferattention}}.

\subsection{Expression recognition results}


This set of experiments makes comparisons between a baseline architecture and the different variants of the proposed architecture. We want to evaluate the relevance of the attention module and the proposed loss function.

{\bf Protocol}. We used distinct metrics to evaluate the proposed methods. Accuracy is calculated as the average number of successes divided by the total number of observations (in this case each face is considered an observation). Precision, recall, F1 score, and confusion matrix are also used in the analysis of the effectiveness of the system. Dem\u{s}ar \cite{demvsar2006statistical} recommends the Friedman test followed by the pairwise Nemenyi test to compare multiple data. The Friedman test is a nonparametric alternative of the analysis of variance (ANOVA) test. The null hypothesis of the test $H_0$ is that all models are equivalent. Similar to the methods in \cite{ptucha2013manifold}, Leave-10-subject-out (L-10-SO) cross-validation was adopted in the evaluation. 

{\bf Results}. Tables \ref{tab:bu3dfe_model_real} and \ref{tab:bu3dfe_model_synth} show the mean and standard deviation for the results obtained on the real and synthetic BU3DFE datasets. The Friedman nonparametric ANOVA test reveals significant differences ($p=0.0498$) between the methods. The Nemenyi post-hoc test was applied to determine which method present significant differences. The result for the Nemenyi post-hoc test (two-tailed test) shows that there are significant differences between the FERAtt+Cls+Rep and all the others, for a significance level at $\alpha < 0.05$.

\begin{table}
\begin{center}
\begin{tabular}{|l|c|c|c|c|c|}
\hline
Method & Acc. & Prec. & Rec. & F1 \\
\hline\hline
PreActResNet18  & 69.37      & 71.48       & 69.56     & 70.50 \\
                & $\pm$2.84  & $\pm$1.46   & $\pm$2.76 & $\pm$2.05  \\
FERAtt+Cls      & 75.15      & 77.34       & 75.45     & 76.38 \\
                & $\pm$3.13  & $\pm$1.40   & $\pm$2.57 & $\pm$1.98  \\
FERAtt+Rep+Cls  & \textbf{77.90}  & \textbf{79.58}  & \textbf{78.05}  & \textbf{78.81} \\
                & $\pm$2.59  & $\pm$1.77   & $\pm$2.34 & $\pm$2.01  \\
\hline
\end{tabular}
\end{center}
\caption{Classification results for the Synthetic BU-3DFE database applied to seven expressions.}
\label{tab:bu3dfe_model_real}
\end{table}

\begin{table}
\begin{center}
\begin{tabular}{|l|c|c|c|c|c|}
\hline
Method & Acc. & Prec. & Rec. & F1 \\
\hline\hline
PreActResNet18  & 75.22& 77.58& 75.49& 76.52\\
                & $\pm$4.60& $\pm$3.72& $\pm$4.68& $\pm$4.19\\
FERAtt+Cls      & 80.41& 82.30& 80.79& 81.54\\
                & $\pm$4.33& $\pm$2.99& $\pm$3.75& $\pm$3.38\\
FERAtt+Rep+Cls  & \textbf{82.11}& \textbf{83.72}& \textbf{82.42}& \textbf{83.06}\\
                & $\pm$4.39& $\pm$3.09& $\pm$4.08& $\pm$3.59\\
\hline
\end{tabular}
\end{center}
\caption{Classification results for the Real BU-3DFE database applied to seven expressions.}
\label{tab:bu3dfe_model_synth}
\end{table}

We repeated the experiment for the Synthetic CK+ dataset and Real CK+ dataset. Tables \ref{tab:ck_model_real} and \ref{tab:ck_model_synth} show the mean and standard deviation for the obtained results. The Friedman  test found significant differences between the methods with a level of significance of $p=0.0388$ for the Synthetic CK+ dataset and $p=0.0381$ for Real CK+ dataset. In this case we applied the Bonferroni-Dunn post-hoc test (one-tailed test) to strengthen the power of the hypotheses test. For a significance level of 0.05, the Bonferroni-Dunn post-hoc test did not show significant differences between the FERAtt+Cls and the Baseline for Synthetic CK+ with $p=0.0216$. When considering FERAtt+Rep+Cls and Baseline methods, it shows significant differences for the Real CK+ dataset with $p=0.0133$.

\begin{table}
\begin{center}
\begin{tabular}{|l|c|c|c|c|c|}
\hline
Method & Acc. & Prec. & Rec. & F1 \\
\hline\hline
PreActResNet18  & 77.63& 68.42& 68.56& 68.49  \\
                & $\pm$2.11& $\pm$2.97& $\pm$1.91& $\pm$2.43      \\
FERAtt+Cls      & 84.60& 74.94& 76.30& 75.61  \\
                & $\pm$0.93& $\pm$0.38& $\pm$1.19& $\pm$0.76      \\
FERAtt+Rep+Cls  & \textbf{85.15}& \textbf{74.68}& \textbf{77.45}& \textbf{76.04} \\
                & $\pm$1.07& $\pm$1.37& $\pm$0.55& $\pm$0.97 \\
\hline
\end{tabular}
\end{center}
\caption{Classification results for the Synthetic CK+ database applied to eight expressions.}
\label{tab:ck_model_real}
\end{table}

\begin{table}
\begin{center}
\begin{tabular}{|l|c|c|c|c|c|}
\hline
Method & Acc. & Prec. & Rec. & F1 \\
\hline\hline
PreActResNet18  & 86.67& 81.62& 80.15& 80.87 \\
                & $\pm$3.15& $\pm$7.76& $\pm$9.50& $\pm$8.63 \\
FERAtt+Cls      & 85.42& 75.65& 78.79& 77.18 \\
                & $\pm$2.89& $\pm$2.77& $\pm$2.30& $\pm$2.55 \\
FERAtt+Rep+Cls  & \textbf{90.30}& \textbf{83.64}& \textbf{84.90}& \textbf{84.25} \\
                & $\pm$1.36& $\pm$5.28& $\pm$8.52& $\pm$6.85 \\
\hline
\end{tabular}
\end{center}
\caption{Classification results for the Real CK+ database applied to eight expressions.}
\label{tab:ck_model_synth}
\end{table}

\begin{figure}[t]
\begin{center}
\includegraphics[width=\linewidth]{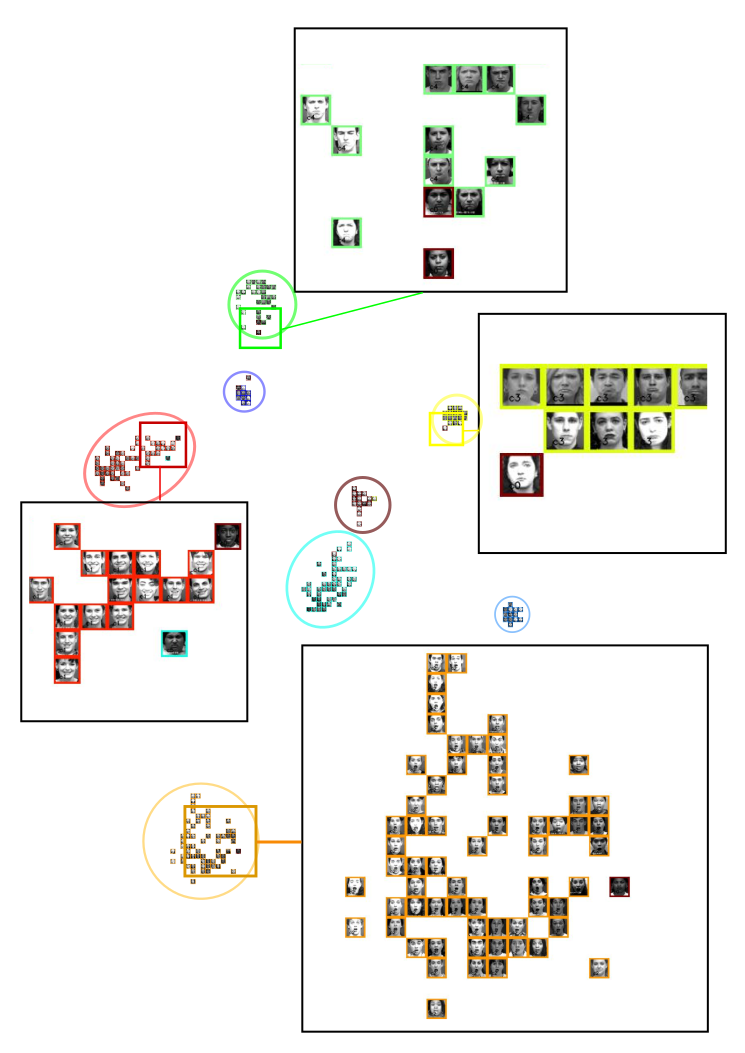}
\end{center}
   \caption{Barnes-Hut t-SNE visualization \cite{van2014accelerating} of the Gaussian Structured loss for the Real CK+ dataset. Each color represents one of the eight emotions including neutral. }
\label{fig:manifold_ck}
\vspace{1mm}
{\color{lightgray}\hrule}
\end{figure}

The results in \figurename~\ref{fig:manifold_ck} show the 64-dimensional embedded space using the Barnes-Hut t-SNE visualization scheme \cite{van2014accelerating} of the Gaussian Structured loss for the Real CK+ dataset. Errors achieved by the network are mostly due to the neutral class which is intrinsically similar to the other expressions we analyzed. Surprisingly, we observed intraclass separations into additional features, such as race, that were not taken into account when modeling or training the network. 
 
\begin{figure*}[t!]
\begin{center}
\begin{tabular}{l}
\includegraphics[width=1.0\linewidth]{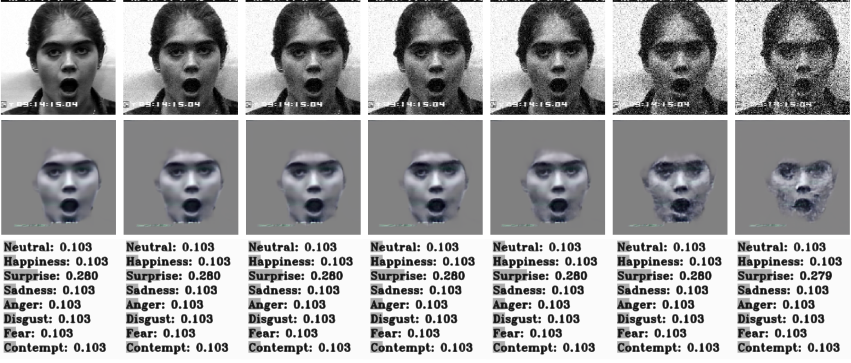} \\
~~~~(a) $\sigma=0.01$ ~~~~~~ (b) $\sigma=0.05$ ~~~~~~ (c) $\sigma=0.07$ ~~~~~~ (d) $\sigma=0.09$ ~~~~~~ (e) $\sigma=0.1$ ~~~~~~~~ (f) $\sigma=0.2$ ~~~~~~~~ (g) $\sigma=0.3$ 
\end{tabular}
\end{center}
   \caption{Attention maps $I_{att}$ under increasing noise levels. We progressively added higher levels of zero mean white Gaussian noise to the same image and tested them using our model. The classification numbers above show the robustness of the proposed approach as the Surprise score and all others are maintained throughout all levels, with a minor change for the highest noise level of 0.30.}
\label{fig:synthetic_noise_result}
\vspace{1mm}
{\color{lightgray}\hrule}
\end{figure*}

\begin{figure}[t]
\begin{center}
\includegraphics[width=\linewidth]{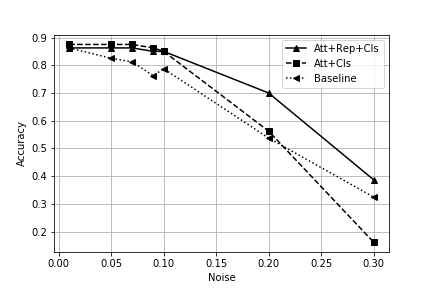}
\end{center}
   \caption{Classification accuracy after adding incremental noise on the Real CK+ dataset. Our approach results in higher accuracy when compared to the baseline, specially for stronger noise levels. Our representation model clearly leverages results showing its importance for classification. Plotted values are the average results for all 325 images in the database.}
\label{fig:noise_real}
\vspace{1mm}
{\color{lightgray}\hrule}
\end{figure}

\begin{figure}[t]
\begin{center}
\includegraphics[width=\linewidth]{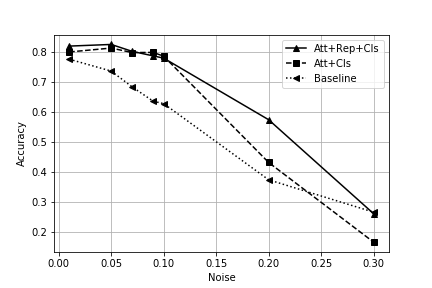}
\end{center}
   \caption{Average classification accuracy after adding incremental noise on the Synthetic CK+ dataset. The behavior of our method in the synthetic data replicates what we have found for the original Real CK+ database, {\it i.e.}, our method is superior to the baseline for all levels of noise. Plotted average values are for 2,000 synthetic images.}
\label{fig:noise_synthetic}
\vspace{1mm}
{\color{lightgray}\hrule}
\end{figure}

\subsection{Robustness to noise}
The objective of this set of experiments is to demonstrate the robustness of our method to the presence of image noise when compared to the baseline architecture PreActResNet18.

{\bf Protocol}.
To carry out this experiment, the Baseline, FERAtt+Class, and FERAtt+Rep+Class models were trained on the Synthetic CK+ dataset. Each of these models was readjusted with increasing noise in the training set ($ \sigma \in [0.05,0.30] $). We maintained the parameters in the training for fine-tuning. We used the real database CK+, and 2000 images were generated for the synthetic dataset for test.

{\bf Results}.
One of the advantages of the proposed approach is that we can evaluate the robustness of the method under different noise levels by visually assessing the changes in the attention map $I_{att}$. \figurename~\ref{fig:synthetic_noise_result} shows the attention maps for an image for white zero mean Gaussian noise levels $\sigma=[ 0.01, 0.05, 0.07, 0.09, 0.1, 0.2, 0.3]$. We observe that our network is quite robust to noise for the range of 0.01 to 0.1 and maintains a distribution of homogeneous intensity values. This aspect is beneficial to the subsequent performance of the classification module. Figures \ref{fig:noise_real} and \ref{fig:noise_synthetic} present classification accuracy results of the evaluated models in the Real CK+ dataset and for 2000 Synthetic images. The proposed method FERAtt+CLs+Rep provides the best classification in both cases.

\section{Conclusions}

In this work, we present a new end-to-end network architecture with an attention model for facial expression recognition. We create a generator of synthetic images which are used for training our models. The results show that, for these experimental conditions, the attention module improves the system classification performance. The loss function presented works as a regularization method on the embedded space contributing positively to the system results. As a future work, we will experiment with larger databases, such as in \cite{BarsoumICMI2016}, which contain images from the real world and are potentially more challenging.

{\small
\bibliographystyle{ieee}
\bibliography{egbib}
}

\end{document}